\documentclass[a4paper,conference]{IEEEtran}

\usepackage{color,amsmath,bm,graphicx,amsfonts,array,booktabs,balance}
\usepackage[compress]{cite}

\ifCLASSINFOpdf
\else

\fi

\begin{document}

\title{Boosting the Discriminant Power of Naive Bayes}

\author{\IEEEauthorblockN{Shihe Wang\IEEEauthorrefmark{1},
Jianfeng Ren\IEEEauthorrefmark{1},
Xiaoyu Lian\IEEEauthorrefmark{2}
Ruibin Bai\IEEEauthorrefmark{1} and
Xudong Jiang\IEEEauthorrefmark{3}
}
\IEEEauthorblockA{\IEEEauthorrefmark{1} School of Computer Science, University of Nottingham Ningbo China\\ 
\IEEEauthorrefmark{2} Ningbo Beilun Taihe High School \\
\IEEEauthorrefmark{3} School of Electrical and Electronic Engineering, Nanyang Technological University}}

\maketitle

\begin{abstract}
Naive Bayes has been widely used in many applications because of its simplicity and ability in handling both numerical data and categorical data. However, lack of modeling of correlations between features limits its performance. 
In addition, noise and outliers in the real-world dataset also greatly degrade the classification performance. 
In this paper, we propose a feature augmentation method employing a stack auto-encoder to reduce the noise in the data and boost the discriminant power of naive Bayes. The proposed stack auto-encoder consists of two auto-encoders for different purposes. 
The first encoder shrinks the initial features to derive a compact feature representation in order to remove the noise and redundant information. The second encoder boosts the discriminant power of the features by expanding them into a higher-dimensional space so that different classes of samples could be better separated in the higher-dimensional space. By integrating the proposed feature augmentation method with the regularized naive Bayes, the discrimination power of the model is greatly enhanced. The proposed method is evaluated on a set of machine-learning benchmark datasets. The experimental results show that the proposed method significantly and consistently outperforms the state-of-the-art naive Bayes classifiers.
\end{abstract}



%
\IEEEpeerreviewmaketitle

\section{Introduction}

Naive Bayes (NB) has been widely used in many applications, e.g., text classification~\cite{hadi2018integrating, kim2018towards, tang2016toward}, action recognition \cite{weng2017spatio}, scene recognition \cite{fornoni2014scene} and malware detection~\cite{yousefi2017autoencoder}. Naive Bayes is a simple and effective classification model. One notable advantage of NB is its ability of handling mixed data types, e.g., both categorical and numerical data. For simplicity, it often assumes that features are independent to each other conditioned on the classification variable. However, the independence assumption rarely holds in reality. 

To address this problem, many approaches have been developed, e.g., structure extension~\cite{jiang2016structure,wu2016sode}, instance selection~\cite{frank2002locally}, instance weighting~\cite{xu2019attribute}, feature selection~\cite{tang2016toward} and feature weighting~\cite{RNB2020shihe,jiang2019class,zhang2021attribute}. Among them, feature weighting approaches~\cite{RNB2020shihe,jiang2019class,zhang2021attribute} have attracted a lot of attention recently, which assign different weights to features to decouple the correlation between features~\cite{RNB2020shihe, jiang2019class,zhang2021attribute}. In~\cite{zhang2021attribute}, attributes and instance are weighted simultaneously. Recently, Wang \emph{et al.} developed a regularized attribute weighting framework to automatically balance the generalization ability and discrimination power of NB classifier~\cite{RNB2020shihe}. These methods partially alleviate the problem, but still not well model the feature correlation.

Artificial defects commonly exist in real-world applications, e.g., missing values or noisy samples. To handle noisy samples and extract the intrinsic data characteristics, many subspace approaches have been developed to remove the unreliable features and extract the discriminant features~\cite{ren2015learning,jiang2008eigenfeature,jiang2008asymmetric,ren2017regularized}. For example, Principal Component Analysis (PCA) is often used for dimensionality reduction by projecting the high-dimensional features into a lower-dimensional space~\cite{jiang2008asymmetric, ren2017regularized}. In literature, auto-encoders have been widely used for filling missing values \cite{choudhury2019imputation} and denoising \cite{vincent2008extracting,yousefi2017autoencoder}.

In this paper, we aim to address the following three challenges of naive Bayes: 1) Removing the noisy and unreliable feature dimensions; 2) Modeling the correlation between features so that the subsequent naive Bayes could make better use of the discriminant information residing in features; 3) Boosting the discriminant power of features. To tackle these three challenges, we resort to stacked auto-encoder \cite{khamparia2020kdsae}.  Stacked auto-encoder is often trained in a self-supervised manner. A portion of the feature entities are intentionally masked off, and the encoder maps the original feature to a lower-dimensional code to remove the noise and uncover the underlying intrinsic data characteristics. The code is then used to reconstruct the original feature~\cite{vincent2008extracting,yousefi2017autoencoder}, with the target of minimizing the reconstruction error. In such a way, the stack auto-encoder could effectively remove the noise, and embed the discriminant information into the compact codes~\cite{hinton2006reducing,yousefi2017autoencoder,vincent2008extracting}. Apparently, the correlation between features is embedded into the codes as well, which is beneficial to the subsequent naive Bayes classifier.

To the best of our knowledge, the stacked auto-encoder has never been used for boosting the discriminant power of features. It is often advantageous to map the feature into a higher-dimensional space so that the features can be linearly separable \cite{webb2003statistical}. The stacked auto-encoder, however, often maps the feature into a compact representation, which many result in discriminant information loss. To tackle this problem, we propose a stacked auto-encoder consisting of two encoders: shrink encoder and expansion encoder. The shrink encoder derives a compact feature representation while the expansion encoder maps the derived compact codes into a higher-dimensional space to enhance the discriminant power of the features. Furthermore, by concatenating the learned representation with the original feature and reconstructed one, the classification performance of the subsequent regularized naive Bayes is significantly improved. 

The proposed Feature-Augmented Regularized Naive Bayes (FAR-NB) is compared with the state-of-the-art NB classifiers on a set of machine-learning datasets for various applications. It significantly and consistently outperforms all the compared methods. The average performance gain on 20 datasets is 5.71\% compared with the second best method, RNB~\cite{RNB2020shihe}.

Our main contributions can be summarized as follows: 1) We propose a feature augmentation method for naive Bayes to exploit the feature correlation and reduce data noise using the stacked auto-encoder. 2) The designed stacked auto-encoder can greatly boost the discriminant power of features by mapping them into a higher-dimensional space, which greatly improves the classification performance. 3) The proposed method is integrated with the regularized naive Bayes and achieves superior performance against state-of-the-art NB classifiers.

\section{Related Works}

\subsection{Naive Bayes Classifiers}
Naive Bayes has been applied in many domains because of its simplicity and noticeable classification performance~\cite{shaban2021accurate,RNB2020shihe,wu2017passive}. Various improved NB classifiers have been developed, which can be broadly divided into five categories: 1) Structure extension methods aim to extend the structure of NB to model the dependency among features~\cite{jiang2016structure}. 2) Instance weighting methods assign different weights to different instances to improve the discrimination power~\cite{xu2019attribute}. 3) Instance selection methods build a local classification model on a subset of training instances to mitigate the effect of noisy samples~\cite{frank2002locally}. 4) Feature selection methods preserve the most representative features by removing the irrelevant or redundant features~\cite{tang2016toward}. 5) Feature weighting approaches weigh the features differently so that the informative feature has a larger weight to enhance the discriminative ability of models~\cite{RNB2020shihe,jiang2019class,zhang2021attribute,lee2011calculating}.


Among these approaches, feature weighting methods achieve a comparably better performance~\cite{RNB2020shihe,jiang2019class,zhang2021attribute}, which can be further divided into filter-based~\cite{zhang2021attribute,lee2011calculating} and wrapper-based approaches~\cite{RNB2020shihe,jiang2019class}. The former utilizes the mutual information~\cite{zhang2021attribute} or KL divergence~\cite{lee2011calculating} to measure the dependency between the feature and the class variable, whereas the latter optimizes the feature weights iteratively by maximizing the classification performance~\cite{RNB2020shihe,jiang2019class}. In AIWNB~\cite{zhang2021attribute}, attribute weights are determined by using the attribute-class relevancy and the average redundancy between each pair of attributes. Zaidi \emph{et al.} developed an attribute weighting model, WANBIA, to derive the attribute weights through a gradient descent optimization procedure~\cite{zaidi2013alleviating}. In CAWNB~\cite{jiang2019class}, different weights are assigned to different features of different classes to enhance the discrimination power of the model. Recently, regularized naive Bayes (RNB) has been developed to automatically balance the class-independent weights and the class-dependent weights~\cite{RNB2020shihe}. These feature weighting approaches emphasize the most discriminative features to improve the classification performance, but fail to model the correlation between features.

\subsection{Feature Extraction Methods}
Feature extraction methods have been widely utilized to discover the compact feature representations from the raw data, which can be broadly categorized into statistical methods~\cite{yang2004two, jiang2008eigenfeature, jiang2008asymmetric} and neural networks~\cite{chen2017nb, yousefi2017autoencoder, kasongo2019deep,khamparia2020kdsae}. The former include Principal Component Analysis~\cite{yang2004two, jiang2008eigenfeature}, Linear Discriminant Analysis~\cite{jiang2008asymmetric} and many others, and the latter include Auto-encoder (AE) \cite{yousefi2017autoencoder,khamparia2020kdsae}, Artificial Neural Network ~\cite{kasongo2019deep}, Convolutional Neural Network (CNN)~\cite{chen2017nb}, and many others.

The auto-encoder encodes the input features in a self-unsupervised way, aiming to derive a compact feature representation by mapping the feature into a lower-dimensional space~\cite{hinton2006reducing}. There are many variations of AEs, e.g., sparse auto-encoder~\cite{hou2019sparse}, denoising auto-encoder~\cite{vincent2008extracting}, contractive auto-encoder~\cite{rifai2011contractive} and convolutional auto-encoder~\cite{zhang2019depth}. In literature, the feature learning approaches for naive Bayes are less explored. In~\cite{yousefi2017autoencoder}, an unsupervised feature learning approach is developed for malware classification using the auto-encoder and the performance of naive Bayes classifier has been greatly improved. Recently, Khamparia \emph{et al.} utilized deep stacked auto-encoder for chronic kidney disease classification to learn representative features~\cite{khamparia2020kdsae}.


\section{Proposed Feature-Augmented Regularized Naive Bayes} 
\subsection{Preliminaries of Regularized Naive Bayes}
\label{sec:preliminaries}

In the Bayesian classification framework, the posterior probability is defined as: 
\begin{equation}
P(c|\bm{x}) = \frac{P(\bm{x}|c)P(c)}{P(\bm{x})} \label{e1},
\end{equation}
where $\bm{x}$ is the feature vector, $c$ is the classification variable, $P(c)$ is the prior probability, $P(\bm{x})$ is the evidence, $P(\bm{x}|c)$ is the likelihood probability distribution and $P(c|\bm{x})$ the posterior probability. Because it is difficult to reliably estimate the likelihood probability $P(\bm{x}|c)$ due to the curse of dimensionality, in naive Bayes methods, the likelihood is often estimated by assuming the feature independence,
\begin{equation}
P(\bm{x}|c) = \prod\limits_{j=1}^mP(x_j|c),
\end{equation}
where $x_j$ is the $j$-th feature dimension of $\bm{x}$ and $m$ is the feature dimensionality. Despite its simplicity, naive Bayes has shown good performance in many applications \cite{hadi2018integrating, kim2018towards, tang2016toward, weng2017spatio, fornoni2014scene, yousefi2017autoencoder}.

Apparently the feature correlation is not modeled in naive Bayes. To address this problem, many feature weighting approaches \cite{RNB2020shihe,jiang2019class,zhang2021attribute} have been developed. In WANBIA \cite{zaidi2013alleviating}, each feature is assigned a different weight to highlight the feature with a large discriminant power,
\begin{equation}
P_I(\bm{x}|c) = \prod\limits_{j=1}^mP(x_j|c)^{\bm{w}_{j}},
\end{equation}
where $\bm{w}_{j}$ is the weight for the $j$-th feature dimension. The weights are optimized by minimizing the mean squared error between the estimated posteriors and the posteriors derived using ground-truth labels. Jiang \emph{et al.} showed that a class-specific weight could further enhance the discrimination power of naive Bayes \cite{jiang2019class},
\begin{equation}
P_D(\bm{x}|c) = \prod\limits_{j=1}^mP(x_j|c)^{\bm{W}_{c,j}} \label{e2},
\end{equation}
where $\bm{W}_{c,j}$ is the entry for the weight matrix $\bm{W}$ for the $j$-th attribute of the class $c$. As a result, different weights are assigned to attributes for different classes. Class-specific attribute weights provide more discriminant power, but the model complexity is considerably increased, so the generalization capability may decrease. To tackle this problem, regularized naive Bayes \cite{RNB2020shihe} determines the likelihood probability as,
\begin{equation}
P_R(\bm{x}|c) = \prod\limits_{j=1}^m \left((1-\alpha)P_D(x_j|c)^{\bm{W}_{c,j}} +\alpha P_I(x_j|c)^{\bm{w}_{j}}\right)
\label{e:post},
\end{equation}
where $P_D(x_j|c)$ is the likelihood weighted using the class-dependent weight matrix $\bm{W}$, $P_I(x_j|c)$ is the likelihood weighted using the class-independent weight vector $\bm{w}$ and $\alpha$ is the hyper-parameter for balancing these two models. The model parameters $\bm{M}=\{\bm{W}, \bm{w}, \alpha\}$ are optimized using a gradient descent procedure \cite{RNB2020shihe}. These weighted naive Bayes \cite{RNB2020shihe,jiang2019class,zhang2021attribute} utilize attribute weights to emphasize the discriminative features. However, they could not fully exploit the discriminant information between features. 

\subsection{Overall Architecture of the Proposed Method}
\begin{figure*}[!ht]
	\centering
	\includegraphics[width=0.85\textwidth]{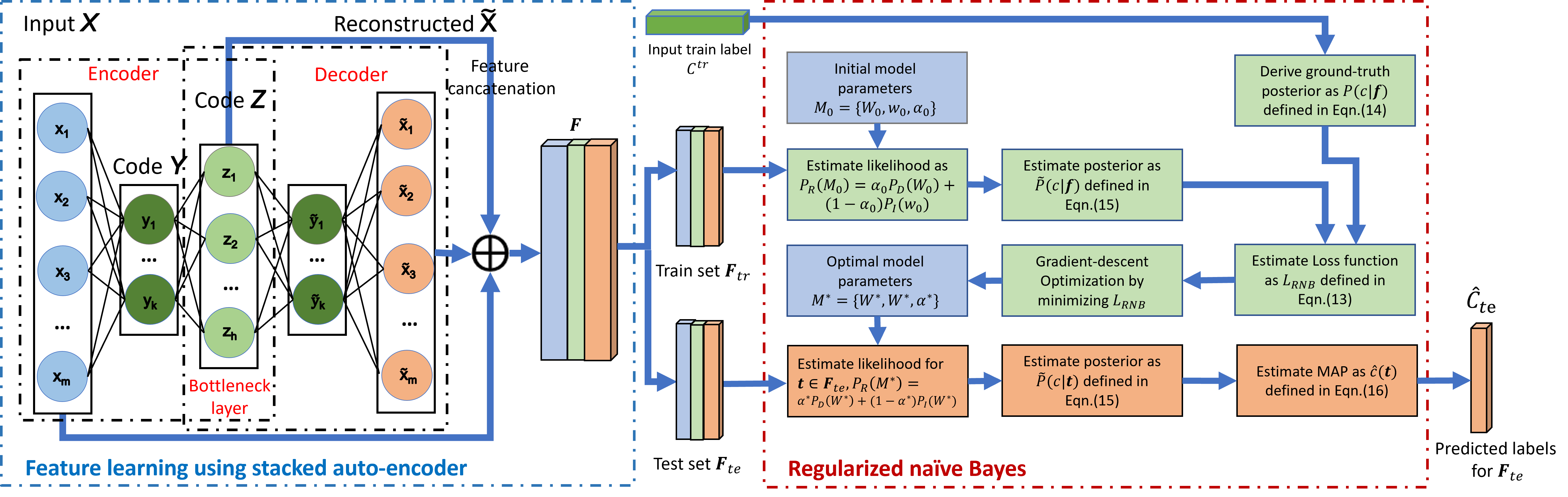}
	\caption{The overall architecture of the proposed method. To remove the noise in data, encode the feature correlation and boost the discriminant power of the model, we propose an stacked auto-encoder to learn a discriminant feature representation. The proposed stacked auto-encoder consists of two encoders: a shrink auto-encoder to derive a compact feature representation to remove noise and unreliable information and an expansion auto-encoder to map the compact code into a higher-dimensional space to boost the discriminant power of the model. The input features $\bm{X}$, the learned feature presentations $\bm{Z}$ and the reconstructed features $\bm{\Tilde{X}}$ are concatenated to form the final feature representation and fed to the subsequent regularized naive Bayes for classification.}
	\label{frame}
\end{figure*}

The proposed method aims to address the following three challenges of previous naive Bayes methods: 1) Noise removal; 2) Encoding the feature correlation; 3) Boosting the discriminant power of naive Bayes. Towards these objectives, we propose a Feature-Augmented Regularized Naive Bayes to learn a discriminant feature representation using an stacked auto-encoder. The overall architecture of the proposed method is shown in Fig.~\ref{frame}. It consists of two main stages: unsupervised feature learning using the stacked auto-encoder and the subsequent regularized naive Bayes. The proposed stacked auto-encoder consists of a shrink encoder to derive the compact feature representation and an expansion encoder to boost the discriminant power of the features. 

Denote the input features as $\bm{X} = \{\bm{x}_1,\bm{x}_2,...,\bm{x}_n\}$, where $\bm{x}_i \in \mathbb{R}^{m}$ is the feature vector for the $i$-th sample, $m$ is the feature dimensionality and $n$ is the number of instances. To remove the noise and encode the correlation information between features, the shrink encoder is designed to learn a compact feature representation $\bm{Y} \in \mathbb{R}^{k\times n}$ using all the initial feature dimensions of $\bm{X}$. Then, the expansion encoder is designed to map $\bm{Y}$ into higher-dimensional features $\bm{Z} \in \mathbb{R}^{h\times n}$ to boost the discriminant power. Then, the reconstructed features $\bm{\Tilde{X}}$ are derived from the codes $\bm{Z}$. The learned features $\bm{Z}$ are concatenated with the original features $\bm{X}$ and the reconstructed ones $\bm{\Tilde{X}}$ as the final features. 

The stacked auto-encoder is trained in a self-supervised way, in which some feature dimensions of $\bm{x}$ are intentionally masked off, and the target is to minimize the reconstruction error, towards the objective of removing the noise in data and unveiling the underlying data characteristics. But different from previous stacked auto-encoders \cite{yousefi2017autoencoder,khamparia2020kdsae} that often derive a compact code from the input feature, in our framework, the stacked auto-encoder is designed to boost the discriminant power of features as well by using the expansion encoder. The number of neurons of the inner layers (feature dimensionality $k$ of $\bm{Y}$ and feature dimensionality $m$ of $\bm{Z}$) of the stacked auto-encoder is automatically adjusted according to optimally remove the data noise and boost the discriminant power. 

Finally, the regularized naive Bayes \cite{RNB2020shihe} is trained using the concatenated features as the input. Some preliminaries of the regularized naive Bayes \cite{RNB2020shihe} are given in Section \ref{sec:preliminaries}. The optimization of the RNB \cite{RNB2020shihe} can be found in Section \ref{sec:opt-RNB}.

\subsection{Feature Learning Using Stacked Auto-encoder}
The designed stacked auto-encoder aims to achieve the following three targets for the subsequent naive Bayes classifier: noise removal, extracting feature correlation and boosting the discriminant power of the model. More specifically, the stacked auto-encoder is designed as a feed-forward network to reconstruct $\bm{X}$ into $\bm{\Tilde{X}}$ with the minimum reconstruction errors. The proposed network contains 
two encoders: shrink encoder and expansion encoder. 

The shrink encoder extracts the intrinsic data characteristics and encodes them into a compact representation, i.e., it maps the input $\bm{X}$ to $\bm{Y} \in \mathbb{R}^{k \times n}$, where $k \leq m$ is the number of neuron in the first inner layers,
\begin{equation}
    \bm{Y} = S(\bm{W}^s\bm{X} + \bm{b}^s ),
\end{equation}
where $S :\mathbb{R}^{m \times n} \rightarrow \mathbb{R}^{k \times n}$ is the activation function of the shrink encoder, $\bm{W}^s \in \mathbb{R}^{k \times m}$ is the weight matrix and $\bm{b}^s \in \mathbb{R}^{k}$ is the bias. The activation function is defined as,
\begin{equation}
\label{eqn:sx}
    S(x) = \begin{cases} 0,& if\ x\leq0, \\x& if\ 0\textless x\leq 1, \\1 &if\ x\geq 1.\end{cases}
\end{equation}

The expansion encoder maps the compact feature $\bm{Y}$ into a higher dimensional space,
\begin{equation}
\bm{Z} = E(\bm{W}^e\bm{Y} + \bm{b}^e),
\end{equation}
where $E :\mathbb{R}^{k \times n} \rightarrow \mathbb{R}^{h \times n}$ is the activation function of the expansion encoder defined similarly as in Eqn. (\ref{eqn:sx}), and $h \geq k$. $\bm{W}^e \in \mathbb{R}^{h \times k}$ is the weight matrix and $\bm{b}^e \in \mathbb{R}^{h}$ is the bias.

During the decoding phase, the encoded feature representations $\bm{Z}$ are transformed back into the original feature space to derive the reconstructed features $\bm{\Tilde{Y}}$,
\begin{equation}
    \bm{\Tilde{Y}} = D(\bm{W}^d\bm{Z} + \bm{b}^d ),
\end{equation}
where the logistic sigmoid function is used for decoding,
\begin{equation}
    D(z) = \frac{1}{1+e^{-z}}.
\end{equation}
Then $\bm{\Tilde{Y}}$ is similarly transformed back to $\bm{\Tilde{X}}$.

The auto-encoder is trained to minimize the Mean Square Error (MSE) between the input $\bm{X}$ and the reconstructed $\bm{\Tilde{X}}$,
\begin{equation}
    L_{AE} = \frac{1}{n}\sum_{i=1}^{n}\sum_{j=1}^{m}(x_{ij} - \Tilde{x}_{ij})^2.
\end{equation}

In the traditional stacked auto-encoder, all the encoders are shrink encoders, aiming to derive a compact feature representation so that the unreliable classification information could be removed and the discriminant information embedded across features can be encoded into $\bm{Z}$. However, some discriminant information may be lost during this process.

\subsection{Boosting Discriminant Power of Regularized Naive Bayes}
To boost the discriminant power of the regularized naive Bayes, we propose to map the compact codes into a higher-dimensional space using the expansion encoder. It remains an open question to determine the optimal feature dimensionalities $k$ of $\bm{Y}$ and $h$ of $\bm{Z}$, as they are affected by many factors. 1) The number of training samples $n$. When $n$ is small, there are insufficient samples to train a reliable network, and hence a smaller network is preferred, i.e., $k$ and $h$ should be kept small. 2) The number of classes. Intuitively, when the number of classes is large, more training samples are needed to reliably estimate the data distribution of each class. Given a fixed number of training samples, we hence prefer a simpler network, i.e., $k$ and $h$ should be smaller. 3) If the input feature dimensionality $m$ is large, there is probably a large amount of redundant information residing in features, and hence we prefer to compress the features into a smaller $k$-dimensional space, and a slightly larger $h$ to boost the discriminant power. 4) If $m$ is relatively small, we prefer to maintain $k$ similar but smaller than $m$ and then map the compact codes into a slightly higher $h$-dimensional space so that the features of different classes are linearly separable. The optimal pair of $(k,h)$ is determined empirically in experiments.  


The learned feature representation $\bm{Z}$, the original features $\bm{X}$ and the reconstructed $\bm{\Tilde{X}}$ all contain discriminant information in different feature spaces. To make full use of all the available discriminant information, we propose to fuse them by concatenating them into the final feature representation as,
\begin{equation}
    \bm{F} = \bm{X} \oplus \bm{Z} \oplus \bm{\Tilde{X}}.
\end{equation}

\subsection{Optimizing Regularized Naive Bayes}
\label{sec:opt-RNB}
The concatenated features $\bm{F}$ are split into the training set $\bm{F}_{tr}$ and the testing set $\bm{F}_{te}$. During the training process, the following loss function is used to optimize the regularized naive Bayes,
\begin{equation}
    L_{RNB} = \frac{1}{2} \sum_{\bm{f_i}\in {\bm{F}}_{tr}}\sum_{c}(P(c|\bm{f}_i)-\Tilde{P}(c|\bm{f}_i))^2, 
    \label{rnb_loss}
\end{equation}
where $P(c|\bm{f}_i)$ is the posterior derived from the ground-truth labels,
\begin{equation}
P(c|\bm{f}_i) = \begin{cases} 1 & if\ c=c_j,\\0&otherwise.\end{cases}
\end{equation}
$\Tilde{P}(c|\bm{f}_i)$ is the estimated posterior with the regularized likelihood function defined in Eqn.~(\ref{e:post}),
\begin{equation}
\Tilde{P}(c|\bm{f}_i) = P(c)P_R(\bm{f}_i|c)	/P(\bm{f}_i).
\label{post_prob}
\end{equation}

The optimal model parameters $\bm{M}^* = \{\alpha^*,\bm{W}^*,\bm{w}^*\}$ of the regularized naive Bayes are derived by minimizing the loss function defined in Eqn.~(\ref{rnb_loss}) using a gradient-descent-based optimization procedure. More details can be found in \cite{RNB2020shihe}.

During testing, the posterior probability $\hat{P}(c|\bm{t})$ for a given testing instance $\bm{t} \in \bm{F}_{te}$ is estimated by using Eqn.~(\ref{post_prob}) with the optimal model $\bm{M}^*$. Finally, the class label for each $\bm{t} \in \bm{F}_{te}$ is derived by using the MAP estimation as follows:
\begin{equation}
    \hat{c}(\bm{t}) = \mathop{\arg\max}_{c \in \bm{C}} \hat{P}(c|\bm{t}),
\end{equation}
where $\bm{C}$ is the set of labels for all classes.

\section{Experimental Results}
\subsection{Experimental Settings}
The proposed FAR-NB is compared with state-of-the-art NB classifiers including RNB~\cite{RNB2020shihe}, WANBIA~\cite{zaidi2013alleviating}, CAWNB~\cite{jiang2019class} and AIWNB~\cite{zhang2021attribute}, as summarized in Table~\ref{des:algo}.
\begin{table}[!ht]
	\centering
	\caption{Summary of compared naive Bayes classifiers.}
	\begin{tabular}{|m{55pt}<{\centering}|m{160pt}|}
		\hline
		\textbf{Algorithm}&\textbf{Description}\\
		\hline
		RNB~\cite{RNB2020shihe}& Wrapper-based regularized attribute weighting method   \\ 
		\hline
		CAWNB~\cite{jiang2019class}&Wrapper-based class-specific attribute weighting method \\
		\hline
		WANBIA~\cite{zaidi2013alleviating}&Wrapper-based class-independent attribute weighting method  \\
		\hline
		AIWNB~\cite{zhang2021attribute}&Filter-based attribute and instance weighting method,  either eager learning AIWNB$^E$ or lazy learning AIWNB$^L$ \\
		\hline
	\end{tabular}
	\label{des:algo}
\end{table}
The experiments are conducted on a collection of benchmark datasets from the University of California at Irvine (UCI) repository~\footnote{https://archive.ics.uci.edu/ml/index.php}, which contains a wide range of domains such as medical, business and biology. The number of instances is distributed between 150 and 10992 and the number of attributes varies between 2 and 60. These 20 machine-learning datasets can provide a comprehensive evaluation of the effectiveness of the proposed method. More details of these datasets are described in Tables~\ref{dataset}. The classification accuracy of each algorithm is derived using 10-fold cross-validation.
\begin{table}[]
\centering
\caption{The datasets are collected from real-world applications in various domains. The number of instances varies between 150 and 10992 and the feature dimensionalities are distributed between 2 and 60.}
\begin{tabular}{@{}ccccc@{}}
\toprule
              & Inst. & Attr. & Class & Domain     \\ \midrule
Balance       & 625   & 4     & 3     & Social     \\
Banana        & 5300  & 2     & 2     & Artificial \\
Banknote      & 1372  & 5     & 2     & Business   \\
Bupa          & 345   & 6     & 2     & Medical    \\
Clevland      & 303   & 13    & 5     & Medical    \\
Contraceptive & 1473  & 9     & 3     & Medical    \\
Ecoli         & 336   & 7     & 8     & Biology    \\
Hayes         & 160   & 4     & 3     & Social     \\
Iris          & 150   & 4     & 3     & Biology    \\
Mammographic  & 961   & 5     & 2     & Medical    \\
Newthyroid    & 215   & 5     & 3     & Medical    \\
Penbased      & 10992 & 16    & 10    & Artificial \\
Satimage      & 6435  & 36    & 7     & Medical    \\
Segment       & 2310  & 19    & 7     & Artificial  \\
Sonar         & 208   & 60    & 2     & Physical   \\
Specfheart    & 267   & 44    & 2     & Physical   \\
Tae           & 151   & 5     & 3     & Education  \\
Vowel         & 990   & 13    & 11    & Artificial  \\
Wine          & 178   & 13    & 3     & Chemical   \\
Yeast         & 1484  & 8     & 10    & Biology    \\ \bottomrule
\end{tabular}
\label{dataset}
\end{table}

\subsection{Ablation Study}
For an ablation study, the proposed method is compared with the following methods:

\noindent \textbf{Original Features}: The original feature is fed into the regularized naive Bayes~\cite{RNB2020shihe} for classification. This comparison could demonstrate the effectiveness of the proposed feature augmentation method in contrast to using the original features.

\noindent \textbf{Baseline}: The stacked auto-encoder \cite{hinton2006reducing} is chosen as the baseline method to derive a compact feature representation and the derived features are fed into the regularized naive Bayes~\cite{RNB2020shihe} for classification. The feature dimension of the bottleneck layer is empirically set to half of the input feature dimensionality. 
The comparison to this baseline can show the power of the proposed feature augmentation method, in contrast to compressing the input feature as in most existing auto-encoders \cite{hinton2006reducing, yousefi2017autoencoder, khamparia2020kdsae}.
\begin{table}[]
	\centering
	\caption{Classification accuracy of the proposed FAR-NB comparing with RNB and baseline method in which the auto-encoder is used to derive a compact feature representation for regularized naive Bayes.}
	\begin{tabular}{@{}cccc@{}}
		\toprule
		& Original Features     & Baseline   & FAR-NB          \\ \midrule
Balance       & 0.7186 & 0.6703 & \textbf{0.8815} \\
Banana        & 0.7338 & 0.4483 & \textbf{0.8621} \\
Banknote      & 0.9278 & 0.8053 & \textbf{0.9854} \\
Bupa          & 0.5327 & 0.5798 & \textbf{0.5882} \\
Clevland      & 0.5773 & 0.5619 & \textbf{0.6237} \\
Contraceptive & 0.5234 & 0.4243 & \textbf{0.5485} \\
Ecoli         & 0.8339 & 0.6640 & \textbf{0.8430} \\
Hayes         & 0.6003 & 0.6101 & \textbf{0.7750} \\
Iris          & 0.9333 & 0.9467 & \textbf{0.9600} \\
Mammographic  & 0.8263 & 0.6671 & \textbf{0.8419} \\
Newthyroid    & 0.9535 & 0.8974 & \textbf{0.9621} \\
Penbased      & 0.9311 & 0.9097 & \textbf{0.9542} \\
Satimage      & 0.8577 & 0.8684 & \textbf{0.8699} \\
Segment       & 0.9459 & 0.8333 & \textbf{0.9593} \\
Sonar         & 0.7742 & 0.6727 & \textbf{0.7983} \\
Specfheart    & 0.8114 & 0.7820 & \textbf{0.8269} \\
Tae           & 0.3440 & 0.3440 & \textbf{0.4683} \\
Vowel         & 0.6465 & 0.5616 & \textbf{0.8192} \\
Wine          & 0.9719 & 0.8595 & \textbf{0.9941} \\
Yeast         & 0.5729 & 0.3982 & \textbf{0.5965} \\ \midrule
    AVG       & 0.7508 & 0.6752 & \textbf{0.8079} \\ \bottomrule
	\end{tabular}
	\label{ablation}
\end{table}

As shown in Table~\ref{ablation}, FAR-NB achieves the highest classification performance on all datasets in comparison to using the original features and the compact feature representation derived using the traditional stacked auto-encoder \cite{hinton2006reducing}. Compared with the original features, the average classification accuracy for the compact features has been greatly reduced by more than 7\%. It shows that directly applying the traditional stacked auto-encoder could not produce good performance. The proposed method utilizes the stacked auto-encoder in a very different way, which greatly boost the discriminant power of the model and hence significantly improves the classification accuracy by 13.27\% on average. These demonstrate the effectiveness of the proposed feature augmentation approach over the traditional stacked auto-encoder. 

\subsection{Comparisons to State-of-the-art Naive Bayes Classifiers}

\begin{table*}[!htpb]
\caption{Classification accuracy for RNB~\cite{RNB2020shihe}, CAWNB~\cite{jiang2019class}, WANBIA~\cite{zaidi2013alleviating}, AIWNB$^E$~\cite{zhang2021attribute}, AIWNB$^L$~\cite{zhang2021attribute} and the proposed FAR-NB. The proposed FAR-NB significantly and consistently outperforms all the compared methods on all the datasets. On average, the performance gain of FAR-NB is 5.71\% compared with the previous best method, RNB~\cite{RNB2020shihe}.}
\centering
\begin{tabular}{@{}ccccccc@{}}
\toprule
              & FAR-NB           & RNB~\cite{RNB2020shihe}    & CAWNB~\cite{jiang2019class}           & WANBIA~\cite{zaidi2013alleviating} & AIWNB$^E$~\cite{zhang2021attribute} & AIWNB$^L$~\cite{zhang2021attribute} \\ \midrule
Balance       & \textbf{0.8815} & 0.7186 & 0.7186 & 0.7186 & 0.7153  & 0.7008  \\
Banana        & \textbf{0.8621} & 0.7338 & 0.7338 & 0.7283 & 0.7198  & 0.7332  \\
Banknote      & \textbf{0.9854} & 0.9278 & 0.9278 & 0.9213 & 0.9206  & 0.9257  \\
Bupa          & \textbf{0.5882} & 0.5327 & 0.5327 & 0.5327 & 0.4202  & 0.4202  \\
Clevland      & \textbf{0.6237} & 0.5773 & 0.5845 & 0.5773 & 0.5717  & 0.5815  \\
Contraceptive & \textbf{0.5485} & 0.5234 & 0.5179 & 0.5139 & 0.5072  & 0.5112  \\
Ecoli         & \textbf{0.8430} & 0.8339 & 0.8338 & 0.8251 & 0.8223  & 0.8223  \\
Hayes         & \textbf{0.7750} & 0.6003 & 0.6003 & 0.6003 & 0.6003  & 0.6003  \\
Iris          & \textbf{0.9600} & 0.9333 & 0.9333 & 0.9333 & 0.9267  & 0.9267  \\
Mammographic  & \textbf{0.8419} & 0.8263 & 0.8252 & 0.8252 & 0.8242  & 0.8232  \\
Newthyroid    & \textbf{0.9621} & 0.9535 & 0.9535 & 0.9580 & 0.9576  & 0.9532  \\
Penbased      & \textbf{0.9542} & 0.9311 & 0.9289 & 0.8988 & 0.8882  & 0.9360  \\
Satimage      & \textbf{0.8699} & 0.8577 & 0.8420 & 0.8440 & 0.8140  & 0.8544  \\
Segment       & \textbf{0.9593} & 0.9459 & 0.9381 & 0.9472 & 0.9264  & 0.9420  \\
Sonar         & \textbf{0.7983} & 0.7742 & 0.7699 & 0.7837 & 0.7649  & 0.7697  \\
Specfheart    & \textbf{0.8269} & 0.8114 & 0.7856 & 0.7854 & 0.7507  & 0.7507  \\
Tae           & \textbf{0.4683} & 0.3440 & 0.3440 & 0.3440 & 0.3244  & 0.3244  \\
Vowel         & \textbf{0.8192} & 0.6465 & 0.6364 & 0.6414 & 0.6364  & 0.6687  \\
Wine          & \textbf{0.9941} & 0.9719 & 0.9719 & 0.9830 & 0.9771  & 0.9660  \\
Yeast         & \textbf{0.5965} & 0.5729 & 0.5756 & 0.5675 & 0.5715  & 0.5715  \\ \midrule
    AVG          & 0.8079          & 0.7508 & 0.7477 & 0.7464 & 0.7320  & 0.7391  \\ 
   W/T/L &       -&12/8/0             &13/7/0 &13/7/0    &15/5/0     &15/5/0      \\ \bottomrule
\end{tabular}
\label{res}
\end{table*}

The comparisons to the state-of-the-art NB methods on 20 benchmark datasets are summarized in Table~\ref{res}. The average classification accuracy of each algorithm over the datasets is summarized at the bottom of Table~\ref{res}, which provides a straightforward comparison of different approaches. To measure the significance of the performance gain, a paired one-tailed t-test with $p=0.05$ significance level is deployed. $W/T/L$ values over all datasets are presented at the bottom of Table~\ref{res}, indicating that the proposed method wins on $W$ datasets, ties on $T$ datasets and loses on $L$ datasets.

As shown in Table~\ref{res}, the proposed FAR-NB consistently outperforms all the compared methods on all the datasets. Among them, FAR-NB is significantly better than RNB, CAWNB, WANBIA, AIWNB$^E$ and AIWNB$^L$ on 12, 13, 13, 15 and 15 datasets, respectively. Compared with wrapper-based attribute weighting methods, e.g. RNB, CAWNB and WANBIA, the proposed FAR-NB obtains the performance gain of 5.71\%, 6.02\% and 6.15\% on average, respectively. Compared with filter-based AIWNB$^E$ and AIWNB$^L$, FAR-NB achieves improvements of 7.59\% and 6.88\% for the average classification accuracy over 20 datasets. These demonstrate the effectiveness of the proposed feature augmentation method.

\begin{figure}[!htpb]
\centering
\includegraphics[width=0.48\textwidth]{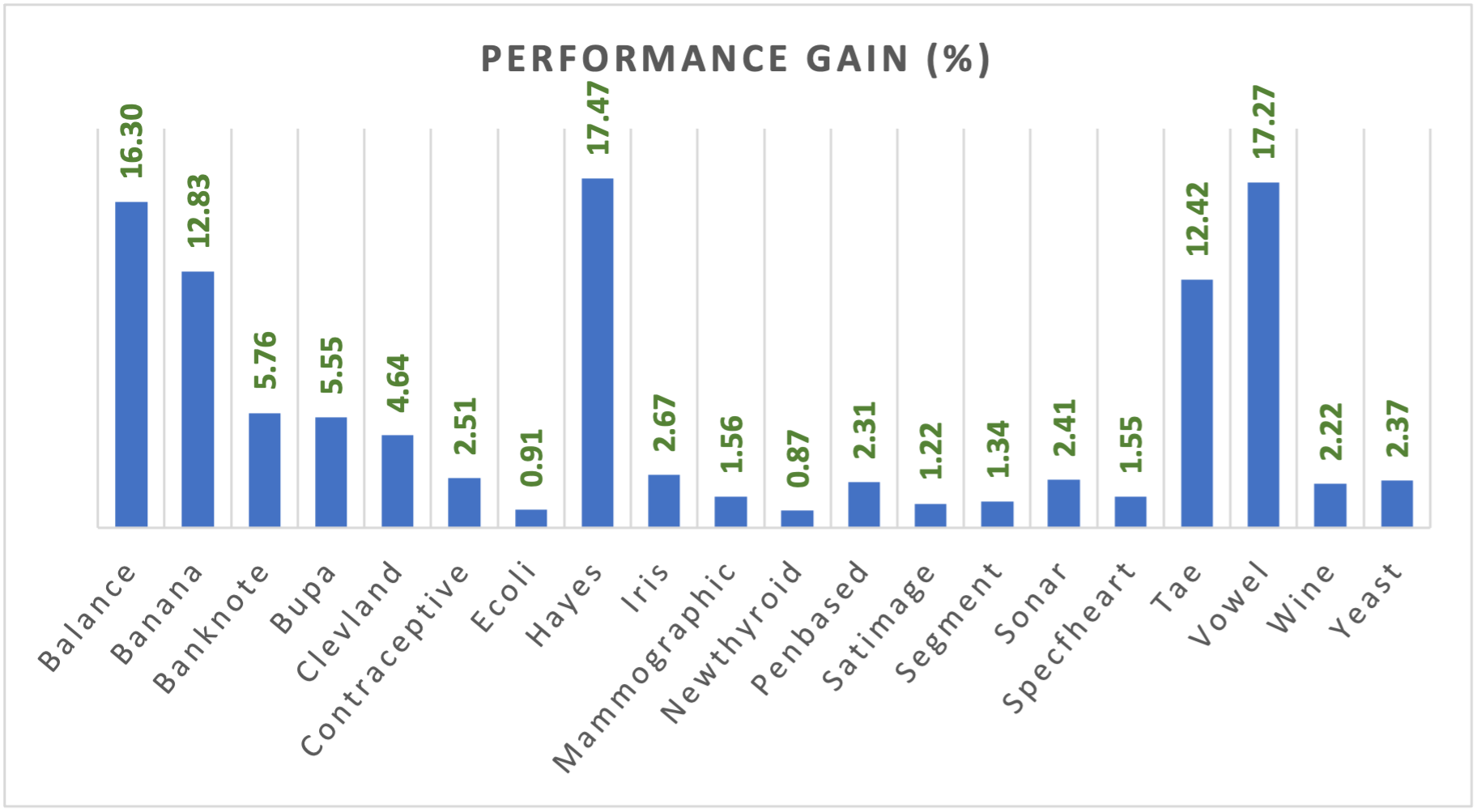}
\caption{The performance gain of the proposed FAR-NB on each dataset compared to RNB~\cite{RNB2020shihe}.}
\label{pg}
\end{figure}

For a better visualization, the performance gain of FAR-NB over the second best performed method, RNB \cite{RNB2020shihe}, on each dataset is shown in Fig.~\ref{pg}. FAR-NB obtains more than 10\% of improvement for classification accuracy on 5 datasets, e.g., `Banana', `Balance', `Hayes', `Tae' and `Vowel'. Besides, FAR-NB can achieve more than 2\% of performance gain compared with RNB \cite{RNB2020shihe} on most datasets.

\section{Conclusion}
The performance of naive Bayes is often limited by lack of the correlation information between features. Many approaches have been developed to alleviate this problem, e.g., feature weighting methods. But these approaches could not fully exploit the discriminant information between features. In this paper, we propose a feature augmentation method for the regularized naive Bayes to extract the discriminant information between features, reduce data noise and boost the discriminant power of the model. Towards these objectives, we resort to the stacked auto-encoder. Different from traditional stacked auto-encoders that map the original features into compact codes, the proposed FAR-NB consists of two encoders, one removes the noise and unreliable information, and another maps the derived compact code into a higher-dimensional space to boost the discriminant power of the model. To further boost the classification performance, the derived features are concatenated with the original features and the reconstructed ones as the augmented features. The proposed feature augmentation method is integrated with the regularized naive Bayes. It is compared with state-of-the-art NB classifiers on 20 datasets for various applications. Experimental results demonstrate that the proposed FAR-NB consistently and significantly outperforms all the compared NB classifiers on all datasets.
\section{Acknowledgment}
This work was supported in part by the National Natural Science Foundation of China under Grant 72071116, and in part by the Ningbo Municipal Bureau Science and Technology under Grants 2019B10026.

\balance
\bibliographystyle{IEEEtran}
\bibliography{IEEEabrv}

\begin{thebibliography}{10}
\providecommand{\url}[1]{#1}
\csname url@samestyle\endcsname
\providecommand{\newblock}{\relax}
\providecommand{\bibinfo}[2]{#2}
\providecommand{\BIBentrySTDinterwordspacing}{\spaceskip=0pt\relax}
\providecommand{\BIBentryALTinterwordstretchfactor}{4}
\providecommand{\BIBentryALTinterwordspacing}{\spaceskip=\fontdimen2\font plus
\BIBentryALTinterwordstretchfactor\fontdimen3\font minus
  \fontdimen4\font\relax}
\providecommand{\BIBforeignlanguage}[2]{{%
\expandafter\ifx\csname l@#1\endcsname\relax
\typeout{** WARNING: IEEEtran.bst: No hyphenation pattern has been}%
\typeout{** loaded for the language `#1'. Using the pattern for}%
\typeout{** the default language instead.}%
\else
\language=\csname l@#1\endcsname
\fi
#2}}
\providecommand{\BIBdecl}{\relax}
\BIBdecl

\bibitem{hadi2018integrating}
W.~Hadi, Q.~A. Al-Radaideh, and S.~Alhawari, ``Integrating associative
  rule-based classification with na{\"\i}ve {Bayes} for text classification,''
  \emph{Applied Soft Computing}, vol.~69, pp. 344--356, 2018.

\bibitem{kim2018towards}
H.-j. Kim, J.~Kim, J.~Kim, and P.~Lim, ``Towards perfect text classification
  with {Wikipedia}-based semantic na{\"\i}ve {Bayes} learning,''
  \emph{Neurocomputing}, vol. 315, pp. 128--134, 2018.

\bibitem{tang2016toward}
B.~Tang, S.~Kay, and H.~He, ``Toward optimal feature selection in naive {Bayes}
  for text categorization,'' \emph{IEEE Transactions on Knowledge and Data
  Engineering}, vol.~28, no.~9, pp. 2508--2521, 2016.

\bibitem{weng2017spatio}
J.~Weng, C.~Weng, and J.~Yuan, ``Spatio-temporal naive-{Bayes} nearest-neighbor
  {(ST-NBNN)} for skeleton-based action recognition,'' in \emph{Proceedings of
  the IEEE Conference on Computer Vision and Pattern Recognition}, 2017, pp.
  4171--4180.

\bibitem{fornoni2014scene}
M.~Fornoni and B.~Caputo, ``Scene recognition with naive {Bayes} non-linear
  learning,'' in \emph{2014 22nd International Conference on Pattern
  Recognition}.\hskip 1em plus 0.5em minus 0.4em\relax IEEE, 2014, pp.
  3404--3409.

\bibitem{yousefi2017autoencoder}
M.~Yousefi-Azar, V.~Varadharajan, L.~Hamey, and U.~Tupakula,
  ``Autoencoder-based feature learning for cyber security applications,'' in
  \emph{2017 International Joint Conference on Neural Networks (IJCNN)}.\hskip
  1em plus 0.5em minus 0.4em\relax IEEE, 2017, pp. 3854--3861.

\bibitem{jiang2016structure}
L.~Jiang, S.~Wang, C.~Li, and L.~Zhang, ``{Structure extended multinomial naive
  Bayes},'' \emph{Information Sciences}, vol. 329, pp. 346--356, 2016.

\bibitem{wu2016sode}
J.~Wu, S.~Pan, X.~Zhu, P.~Zhang, and C.~Zhang, ``Sode: Self-adaptive
  one-dependence estimators for classification,'' \emph{Pattern Recognition},
  vol.~51, pp. 358--377, 2016.

\bibitem{frank2002locally}
E.~Frank, M.~Hall, and B.~Pfahringer, ``Locally weighted naive {Bayes},'' in
  \emph{Proceedings of the Nineteenth conference on Uncertainty in Artificial
  Intelligence}, 2002, pp. 249--256.

\bibitem{xu2019attribute}
W.~Xu, L.~Jiang, and L.~Yu, ``An attribute value frequency-based instance
  weighting filter for naive {Bayes},'' \emph{Journal of Experimental \&
  Theoretical Artificial Intelligence}, vol.~31, no.~2, pp. 225--236, 2019.

\bibitem{RNB2020shihe}
S.~{Wang}, J.~{Ren}, and R.~{Bai}, ``A regularized attribute weighting
  framework for naive {Bayes},'' \emph{IEEE Access}, vol.~8, pp.
  225\,639--225\,649, 2020.

\bibitem{jiang2019class}
L.~Jiang, L.~Zhang, L.~Yu, and D.~Wang, ``{Class-specific attribute weighted
  naive Bayes},'' \emph{Pattern Recognition}, vol.~88, pp. 321--330, 2019.

\bibitem{zhang2021attribute}
H.~Zhang, L.~Jiang, and L.~Yu, ``Attribute and instance weighted naive
  {Bayes},'' \emph{Pattern Recognition}, vol. 111, p. 107674, 2021.

\bibitem{ren2015learning}
J.~Ren, X.~Jiang, and J.~Yuan, ``Learning {LBP} structure by maximizing the
  conditional mutual information,'' \emph{Pattern Recognition}, vol.~48,
  no.~10, pp. 3180--3190, 2015.

\bibitem{jiang2008eigenfeature}
X.~Jiang, B.~Mandal, and A.~Kot, ``Eigenfeature regularization and extraction
  in face recognition,'' \emph{IEEE Transactions on Pattern Analysis and
  Machine Intelligence}, vol.~30, no.~3, pp. 383--394, 2008.

\bibitem{jiang2008asymmetric}
X.~Jiang, ``Asymmetric principal component and discriminant analyses for
  pattern classification,'' \emph{IEEE Transactions on Pattern Analysis and
  Machine Intelligence}, vol.~31, no.~5, pp. 931--937, 2008.

\bibitem{ren2017regularized}
J.~Ren and X.~Jiang, ``Regularized {2-D} complex-log spectral analysis and
  subspace reliability analysis of {micro-Doppler} signature for {UAV}
  detection,'' \emph{Pattern Recognition}, vol.~69, pp. 225--237, 2017.

\bibitem{choudhury2019imputation}
S.~J. Choudhury and N.~R. Pal, ``Imputation of missing data with neural
  networks for classification,'' \emph{Knowledge-Based Systems}, vol. 182, p.
  104838, 2019.

\bibitem{vincent2008extracting}
P.~Vincent, H.~Larochelle, Y.~Bengio, and P.-A. Manzagol, ``Extracting and
  composing robust features with denoising autoencoders,'' in \emph{Proceedings
  of the 25th International Conference on Machine learning}, 2008, pp.
  1096--1103.

\bibitem{khamparia2020kdsae}
A.~Khamparia, G.~Saini, B.~Pandey, S.~Tiwari, D.~Gupta, and A.~Khanna,
  ``{KDSAE}: Chronic kidney disease classification with multimedia data
  learning using deep stacked autoencoder network,'' \emph{Multimedia Tools and
  Applications}, vol.~79, no.~47, pp. 35\,425--35\,440, 2020.

\bibitem{hinton2006reducing}
G.~E. Hinton and R.~R. Salakhutdinov, ``Reducing the dimensionality of data
  with neural networks,'' \emph{Science}, vol. 313, no. 5786, pp. 504--507,
  2006.

\bibitem{webb2003statistical}
A.~R. Webb, \emph{Statistical pattern recognition}.\hskip 1em plus 0.5em minus
  0.4em\relax John Wiley \& Sons, 2003.

\bibitem{shaban2021accurate}
W.~M. Shaban, A.~H. Rabie, A.~I. Saleh, and M.~Abo-Elsoud, ``Accurate detection
  of {COVID}-19 patients based on distance biased na{\"\i}ve {Bayes} ({DBNB})
  classification strategy,'' \emph{Pattern Recognition}, p. 108110, 2021.

\bibitem{wu2017passive}
Z.~Wu, Q.~Xu, J.~Li, C.~Fu, Q.~Xuan, and Y.~Xiang, ``Passive indoor
  localization based on {CSI} and naive {Bayes} classification,'' \emph{IEEE
  Transactions on Systems, Man, and Cybernetics: Systems}, vol.~48, no.~9, pp.
  1566--1577, 2017.

\bibitem{lee2011calculating}
C.-H. Lee, F.~Gutierrez, and D.~Dou, ``Calculating feature weights in naive
  {Bayes} with {Kullback-Leibler} measure,'' in \emph{2011 IEEE 11th
  International Conference on Data Mining}.\hskip 1em plus 0.5em minus
  0.4em\relax IEEE, 2011, pp. 1146--1151.

\bibitem{zaidi2013alleviating}
N.~A. Zaidi, J.~Cerquides, M.~J. Carman, and G.~I. Webb, ``Alleviating naive
  {Bayes} attribute independence assumption by attribute weighting,'' \emph{The
  Journal of Machine Learning Research}, vol.~14, no.~1, pp. 1947--1988, 2013.

\bibitem{yang2004two}
J.~Yang, D.~Zhang, A.~F. Frangi, and J.-y. Yang, ``Two-dimensional {PCA}: a new
  approach to appearance-based face representation and recognition,''
  \emph{IEEE Transactions on Pattern Analysis and Machine Intelligence},
  vol.~26, no.~1, pp. 131--137, 2004.

\bibitem{chen2017nb}
F.-C. Chen and M.~R. Jahanshahi, ``{NB-CNN}: {Deep} learning-based crack
  detection using convolutional neural network and na{\"\i}ve {Bayes} data
  fusion,'' \emph{IEEE Transactions on Industrial Electronics}, vol.~65, no.~5,
  pp. 4392--4400, 2017.

\bibitem{kasongo2019deep}
S.~M. Kasongo and Y.~Sun, ``A deep learning method with filter based feature
  engineering for wireless intrusion detection system,'' \emph{IEEE Access},
  vol.~7, pp. 38\,597--38\,607, 2019.

\bibitem{hou2019sparse}
L.~Hou, V.~Nguyen, A.~B. Kanevsky, D.~Samaras, T.~M. Kurc, T.~Zhao, R.~R.
  Gupta, Y.~Gao, W.~Chen, D.~Foran \emph{et~al.}, ``Sparse autoencoder for
  unsupervised nucleus detection and representation in histopathology images,''
  \emph{Pattern Recognition}, vol.~86, pp. 188--200, 2019.

\bibitem{rifai2011contractive}
S.~Rifai, P.~Vincent, X.~Muller, X.~Glorot, and Y.~Bengio, ``Contractive
  auto-encoders: explicit invariance during feature extraction,'' in
  \emph{Proceedings of the 28th International Conference on International
  Conference on Machine Learning}, 2011, pp. 833--840.

\bibitem{zhang2019depth}
Z.~Zhang, D.~Chen, Z.~Wang, H.~Li, L.~Bai, and E.~R. Hancock, ``Depth-based
  subgraph convolutional auto-encoder for network representation learning,''
  \emph{Pattern Recognition}, vol.~90, pp. 363--376, 2019.

\end{thebibliography}

\end{document}